# US-Cut: Interactive Algorithm for Rapid Detection and Segmentation of Liver Tumors in Ultrasound Acquisitions

Jan Egger[a,b], Philip Voglreiter[a], Mark Dokter[a], Michael Hofmann[a], Xiaojun Chen[c]
Wolfram G. Zoller[d], Dieter Schmalstieg[a] and Alexander Hann[d]

[a] TU Graz, Institute for Computer Graphics and Vision, Inffeldgasse 16, 8010 Graz, Austria
[b] BioTechMed, Krenngasse 37/1, 8010 Graz, Austria
[c] Shanghai Jiao Tong University, School of Mechanical Engineering, 800 Dong Chuan Road, Shanghai 200240, China
[d] Dept. of Internal Medicine and Gastroenterology, Katharinenhospital, Kriegsbergstrasse 60, 70174 Stuttgart, Germany

## ABSTRACT

Ultrasound (US) is the most commonly used liver imaging modality worldwide. It plays an important role in follow-up of cancer patients with liver metastases. We present an interactive segmentation approach for liver tumors in US acquisitions. Due to the low image quality and the low contrast between the tumors and the surrounding tissue in US images, the segmentation is very challenging. Thus, the clinical practice still relies on manual measurement and outlining of the tumors in the US images. We target this problem by applying an interactive segmentation algorithm to the US data, allowing the user to get real-time feedback of the segmentation results. The algorithm has been developed and tested hand-in-hand by physicians and computer scientists to make sure a future practical usage in a clinical setting is feasible. To cover typical acquisitions from the clinical routine, the approach has been evaluated with dozens of datasets where the tumors are hyperechoic (brighter), hypoechoic (darker) or isoechoic (similar) in comparison to the surrounding liver tissue. Due to the interactive real-time behavior of the approach, it was possible even in difficult cases to find satisfying segmentations of the tumors within seconds and without parameter settings, and the average tumor deviation was only 1.4mm compared with manual measurements. However, the long term goal is to ease the volumetric acquisition of liver tumors in order to evaluate for treatment response. Additional aim is the registration of intraoperative US images via the interactive segmentations to the patient's pre-interventional CT acquisitions.

**Keywords:** Ultrasound, Liver Tumors, Segmentation, Detection, Interactive, Rapid.

## 1. DESCRIPTION OF PURPOSE

Ultrasound (US) is the most commonly used liver imaging modality worldwide. It is an easily accessible and less expensive imaging procedure compared to computed tomography (CT) or magnetic resonance imaging (MRI). These are two of the main reasons why guidelines included the use of US as one of the first procedures to evaluate for liver metastases at the time of diagnosing a patient with cancer. Furthermore, US is one of the main imaging procedures to evaluate response to palliative treatment of cancer patients[1]. In a setting of palliative treatment upon a determined number of chemotherapy cycles, evaluation of the size of liver metastases is necessary and can change the management of patients. An increase in size defines a progressive disease, requiring the alteration of the management, like a change of chemotherapy regimen. A shrinkage or constant size often results in continuation of the chemotherapy regimen. The appearance of liver metastasis in US is highly variable[2]. Figure 1 gives a schematic overview of the different echo pattern. In comparison to the liver tissue, metastases may appear hyper- or hypoechoic (brighter (A) or darker (C)) in B-mode. Almost isoechoic (B) masses with a similar echo pattern as the surrounding liver tissue are challenging to find, and determination of their size is difficult. A hypoechoic halo (darker rim around the liver lesion) is often seen around hyper (D) or isoechoic (E) metastasis. These multiple appearances of liver lesions including additional morphologic changes over time contribute to one of the main issues of US, the observer dependence[3]. A novel easily applicable tool to automatically measure the size of a metastasis with different echogenic pattern is the first step in reliably evaluating the growth of liver masses over time. It may reduce some of the issues with the poor inter-observer agreement of US images.

______________________
E-mail: jan.egger@icg.tugraz.at

Others working in the field of liver tumor segmentation in US images are Boen[4], Pradeep Kumar et al.[5], Yoshida et al.[6], and Poonguzhali and Ravindran[7]. Boen presents a method called seeded region growing (SRG), which isolates regions of interest for later processing. Kumar et al. introduce a fully automatic segmentation approach, which uses statistical features to help distinguishing between normal and ultrasonic tumor liver images. Yoshida et al. propose a method for the segmentation of low-contrast objects embedded in noisy images. Poonguzhali and Ravindran introduce a complete automatic technique for the segmentation of masses in ultrasound images by using a region growing approach. In this study, however, we present an interactive real-time segmentation method for liver tumors in ultrasound images. Hence, our very fast interactive segmentation approach, can also be applied intraoperatively during an intervention, in contrast to the existing approaches. The approach is easy and intuitive to use and allows a rapid segmentation by moving the mouse over the image and stopping when a satisficing segmentation is achieved. To enable this interactive real-time segmentation, a special graph-based algorithm was designed, that allows a very fast calculation of an s-t-cut[8] and needs only one user-defined seed point to set up the graph. However, even if there exist semi-automatic approaches for the segmentation of tumors in US images, the authors are not aware of an interactive real-time segmentation approach in this research area.

This contribution is organized as follows: Section 2 presents details of the methods, Section 3 discusses experimental results and Section 4 concludes the paper and gives an outlook on future work.

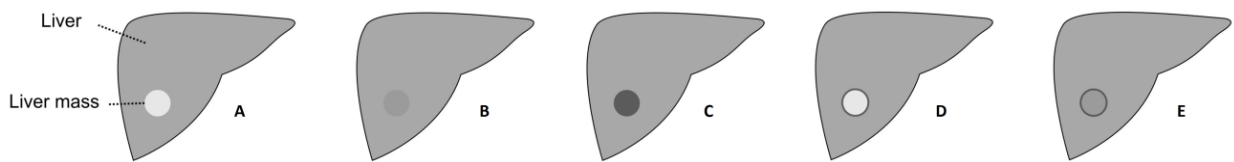

Fig. 1: Schematic overview of the different echo pattern of homogenous liver masses in ultrasound B-mode.

## 2. METHODS

The algorithm presented here belongs to the class of graph-based approaches[9-13], where an image is interpreted as a graph $G(V,E)$, consisting of nodes $n \in V$ and edges $e \in E$. The nodes are sampled in the image, plus two virtual nodes $s \in V$ (called the source) and $t \in V$ (called the sink) to calculate an s-t-cut after the graph construction. The set of edges establish the connections between the (virtual) nodes. The edge $\langle v_i, v_j \rangle \in E$ connects the two nodes $v_i, v_j$ [14-19]. In a first step, the nodes for the graph are sampled along radial lines (rays) distributed equidistantly clockwise around a fixed point inside the image (left image of Figure 2). So-called $\infty$-weighted intra-edges are constructed, which connect nodes along the same ray (Figure 2, second image from the left). These intra-edges ensure that the s-t-cut uses exact one edge out of a set of edges that belong to the same ray. Then, the $\infty$-weighted inter-edges are constructed (two rightmost images of Figure 2), which connect nodes from different rays under the delta value Δr, which influences the number of possible s-t-cuts and therefore the flexibility or smoothness of the resulting segmentation (with $r = (0,...,R-1)$ and $R$ as the number of rays). Finally, to complete the graph, weighted edges between the nodes and the source $s$ and the sink $t$ are constructed. The weights of these edges depend on the gray values in the image. The graph construction relates to previous works[20-23], however, instead of templates with fixed seed point positions, we used a circular template, which can be interactively positioned[24-26] by the user on the image. This is the first time we applied the approach to ultrasound data.

The overall workflow of the presented approach for interactive detection and segmentation of liver tumors is presented in Figure 3. The left image of Figure 3 shows a circular template that is used for the underlying graph. The second image of Figure 3 represents the principle of the graph construction that is based on the underlying circular template. Rays are sent out radially from the center point of the circle and along these rays, the nodes for the graph are sampled. The inter- and intra-edges are constructed between the nodes, whereby the construction of the inter-edges depends on the delta value Δr. The third image of Figure 3 shows a complete graph (green) that is constructed in an ultrasound image (note: in general, the graph is not displayed to the user, rather the segmentation result is directly shown). The position of the graph depends on the interactive placement of the mouse cursor (center point of the circle/graph) by the user. The rightmost image of Figure 3 presents the segmentation result (red) displayed to the user based on the current center point of the circle/graph, the seed point (white).

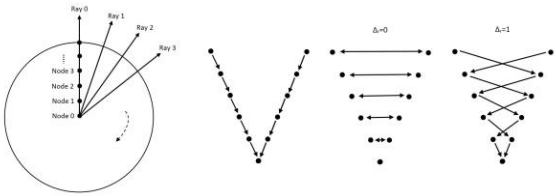

Fig. 2: Principle of the graph construction that is used for the interactive segmentation process.

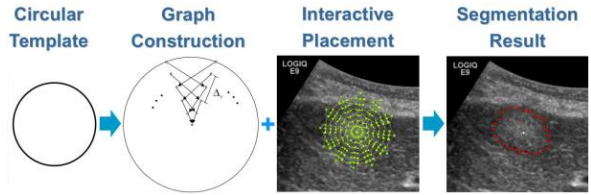

Fig. 3: Overall workflow of the presented approach for interactive detection and segmentation of liver tumors.

## 3. RESULTS

Ultrasound examinations were performed using a multifrequency curved probe, which allows ultrasound acquisitions with a bandwith of 1 to 6 MHz (LOGIQ E9/GE Healthcare, Milwaukee, Il, USA, and Toshiba Aplio 80, Otawara, Japan). Using the digital picture archive of our ultrasound unit we retrospectively selected images of liver masses examined by Dr. Hann in the year 2013 that fulfilled the following criterion to ease the segmentation: The main criterion consisted of the echogenicity class of the lesion (hypo-, hyper and isoechoic). During the selection process, we excluded cystic or marked calcified masses, so that the chosen lesions appeared relatively homogenous. Pictures with visible marker or text overlaying the target lesion were excluded. After identifying one lesion per echogenicity class, the selection process was stopped, patient information was removed from the image and the anonymized picture was subsequently processed by the segmentation algorithm. For evaluation algorithmic segmentations have been compared with the manual segmentations from a physician (Table 1), which lead to an average deviation of 1.4 mm.

| Case   | 1     | 2     | 3     | 4     | 5     | 6     | 7     | 8     | 9    | 10    | $\mu \pm \sigma$ |
|--------|-------|-------|-------|-------|-------|-------|-------|-------|------|-------|------------------|
| Manual | 11.32 | 18.5  | 23.92 | 13.95 | 12.1  | 30.49 | 15.63 | 21.66 | 7.18 | 19.83 | 17.46±6.86       |
| US-Cut | 10.66 | 17.22 | 21.94 | 10.68 | 10.78 | 28.03 | 12.7  | 21.88 | 6.77 | 19.3  | 16.03±6.62       |

Tab. 1: Manual vs algorithm segmentations of maximum tumor diameter for ten cases (in mm).

Figure 4 presents the interactive segmentation results (red dots) of several liver tumors, a metastasis of a neuroendocrine neoplasm of the pancreas (upper left), a metastasis of a colon cancer (upper right) and two different views of a metastasis of an uveal melanoma (lower images). The tumor in the upper left image appears hyperechoic compared to the surrounding liver tissue, while the tumors in the lower images appear hypoechoic than the surrounding structures. However, the tumor of the upper right image has dark and bright areas. The white dots represent the seed points, where the user was satisfied with the automatic outlining of the tumors and therefore stopped the interactive segmentation process at this position.

Figure 5 presents a manual and an interactive segmentation result of a tumor which appears hyperechoic (brighter) with a hypoechoic (darker) halo (metastasis of a colon cancer) in comparison to the surrounding liver tissue. The tumor exhibits a very low contrast to the surrounding liver parenchyma. The left side of Figure 5 shows the native image with a zoomed area of the tumor. In the middle of Figure 5, the manual measurement result (white dotted line between two white pluses) of the tumor is shown. Finally, the right side of Figure 5 shows interactive segmentation results (red dots) and seed point at this position (white).

Figure 6 presents a direct side-by-side comparison of the interactively achieved segmentation results (right side) and a manual expert measurement (left side) for two liver metastases. The presented approach was realized in C++ within the medical prototyping platform MeVisLab from Fraunhofer MeVis in Bremen, Germany (www.mevislab.de)[27-31]. In our implementation, the real-time segmentation could be performed on a Macbook Pro laptop computer with an Intel Core i7-2860QM CPU, 4x2.50 GHz, 8 GB RAM, Windows 7 Professional x64.

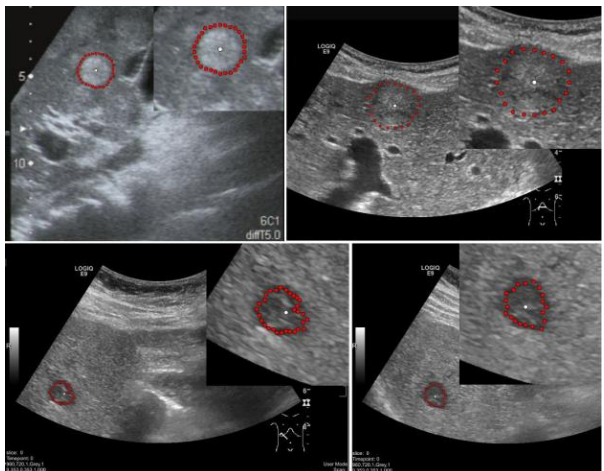

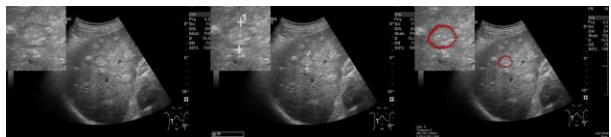

Fig. 5: Manual and interactive segmentation result of a hyperechoic appearing tumor with a hypoechoic halo (colon cancer metastasis).

Fig. 4: Interactive segmentation results (red dots) of several different liver tumors. The white dots represent the seed points where the user was satisfied with the automatic outlining of the tumors and therefore stopped the interactive segmentation process at this position.

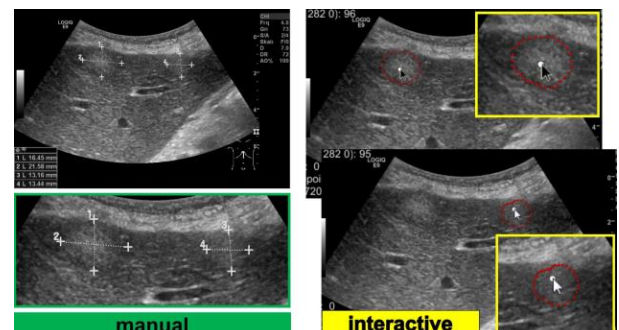

Fig. 6: Side-by-side comparison of manual (left) and interactive segmentation results for the two liver metastases.

## 4. CONCLUSIONS

In this study, we present the results of an interactive segmentation approach for liver tumors in US acquisitions. This is a very challenging problem and still under active research, because of the low image quality and the low contrast between the tumors and the surrounding liver tissue in US images. As a consequence, in clinical practice, the tumors are still outlined and measured purely manual, which leads to poor inter-observer agreement regarding the tumor size. We supported the manual segmentation task by applying an interactive segmentation algorithm to the US data, so the user gets real-time feedback of the segmentation results. To ensure a future practical usage in a clinical setting, the approach was developed and tested hand-in-hand by physicians and computer scientists. To cover typical acquisitions from the clinical routine, the approach was tested with data where the tumors are hyperechoic (brighter), hypoechoic (darker) or have a nearly isoechoic (similar) appearance to the surrounding liver tissue. The approach's interactive real-time performance enabled satisfying segmentations within a few seconds even for difficult cases and without new parameter settings. Limitations of this study include the small number of patients involved and the exclusion of marked inhomogenous, calcified or cystic lesions. In summary, the achieved research highlights of the presented work are: 1. A special interactive segmentation approach has been developed; 2. Interactive behavior allows real-time feedback of the segmentations; 3. The approach is fast, intuitive and user friendly; 4. It enables to find satisfying segmentations within seconds; 5. It was tested with data from the current clinical routine; 6. It worked with different appearances of liver tumors in US acquisitions.

In contrast to the approaches from the introduction[4-7], we focused with this study on a very fast interactive segmentation approach, which can be performed in real-time and thus also be applied intraoperatively during an intervention. These also include liver tumor ablations, where intraoperative US and CT data is acquired that has to be aligned for treatment planning. However, because of the required interactive input from the user, the presented approach is not suitable to process large amount of data sets (e.g., from a database for automatic analysis or classification) without further ado. Existing approaches are more appropriate in this case.

There are several areas for future work: in particular, the evaluation of the method using a greater set of images and comparing the results of different operators and further measurement of liver lesions during treatment over time, including heterogeneous masses. Furthermore, we have started acquiring 3D ultrasound data of patients with liver tumors like Xu and colleagues[32] for a volumetric analysis. However, in contrast to the mentioned study, where the segmentation was performed manually within the software of the manufacture, we plan to adapt and use our interactive segmentation approach, to support this very time-consuming outlining process even in 3D.

## ACKNOWLEDGEMENT

The work received funding from the European Union in FP7: Clinical Intervention Modelling, Planning and Proof for Ablation Cancer Treatment (ClinicIMPPACT, grant agreement no. 610886) and Generic Open-end Simulation Environment for Minimally Invasive Cancer Treatment (GoSmart, grant agreement no. 600641). Dr. Xiaojun Chen receives support from NSFC (National Natural Science Foundation of China) grant 81171429. Dr. Dr. Jan Egger receives funding from BioTechMed-Graz ("Hardware accelerated intelligent medical imaging"). A video demonstrating the interactive real-time segmentation of liver tumors in an Ultrasound image can be found under the following YouTube-channel:

https://www.youtube.com/c/JanEgger/

## REFERENCES

1. Seufferlein, T. et al. and ESMO Guidelines Working Group "Pancreatic adenocarcinoma: ESMO-ESDO Clinical Practice Guidelines for diagnosis, treatment and follow-up," *Ann Oncol.* 23 Suppl 7:vii33-40 (2012).
2. Harvey, C. J. and Albrecht, T. "Ultrasound of focal liver lesions," *Eur Radiol.* (9):1578-93 (2011).
3. Hohmann, J. et al. "Characterisation of focal liver lesions with unenhanced and contrast enhanced low MI real time ultrasound: on-site unblinded versus off-site blinded reading," *Eur J Radiol.* 81(3):e317-24 (2012).
4. Boen, D. "Segmenting 2D Ultrasound Images using Seeded Region Growing," *University of British Columbia*, Department of Electrical and Computer Engineering, Vancouver, British Columbia, pp. 1-11 (2006).
5. Pradeep Kumar, B. P. et al. "An Automatic Approach for Segmentation of Ultrasound Liver Images," *International Journal of Emerging Technology and Advanced Engineering*, Volume 3, Issue 1 (2013).
6. Yoshida, H. et al., "Segmentation of liver tumors in ultrasound images based on scale-space analysis of the continuous wavelet transform," *IEEE Ultrasonics Symposium*, pp. 1713-1716, vol.2 (1998).
7. Poonguzhali, S. and Ravindran, G. A. "A complete automatic region growing method for segmentation of masses on ultrasound images," *International Conference on Biomedical and Pharmaceutical Engineering (ICBPE)*, pp. 88-92 (2006).
8. Boykov, Y. and Kolmogorov, V. "An Experimental Comparison of Min-Cut/ Max-Flow Algorithms for Energy Minimization in Vision," *IEEE Transactions on Pattern Analysis and Machine Intelligence*, 26(9), pp. 1124-1137 (2004).
9. Shi, J. and Malik, J. "Normalized Cuts and Image Segmentation," *IEEE PAMI*, vol. 22, no. 8 (2000).
10. Egger, J. et al. "Nugget-Cut: A Segmentation Scheme for Spherically- and Elliptically-Shaped 3D Objects," *32nd Annual Symposium of the German Association for Pattern Recognition (DAGM),* LNCS 6376, pp. 383–392, Springer Press, Darmstadt, Germany (2010).
11. Egger, J. et al. "Graph-Based Tracking Method for Aortic Thrombus Segmentation," *Proceedings of 4th European Congress for Medical and Biomedical Engineering, Engineering for Health*, Antwerp, Belgium, Springer, pp. 584-587 (2008).
12. Egger, J. et al. "Pituitary Adenoma Segmentation," *In: Proceedings of International Biosignal Processing Conference*, Charité, Berlin, Germany (2010).
13. Bauer, M. et al. "A fast and robust graph-based approach for boundary estimation of fiber bundles relying on fractional anisotropy maps," *20th International Conference on Pattern Recognition (ICPR)*, Istanbul, Turkey, pp. 4016-4019 (2010).
14. Li, K., Wu, X., Chen, D. Z. and Sonka, M. "Optimal Surface Segmentation in Volumetric Images - A Graph-Theoretic Approach," *IEEE Transactions on Pattern Analysis and Machine Intelligence*. 28(1):119-134 (2006).
15. Egger, J. et al. "Aorta Segmentation for Stent Simulation," *12th International Conference on Medical Image Computing and Computer Assisted Intervention (MICCAI), Cardiovascular Interventional Imaging and Biophysical Modelling Workshop*, 10 pages, London, UK (2009).
16. Egger, J. et al "A flexible semi-automatic approach for glioblastoma multiforme segmentation," *Proceedings of International Biosignal Processing Conference*, Charité, Berlin, Germany (2010).
17. Bauer, M. et al. "Boundary estimation of fiber bundles derived from diffusion tensor images," *International journal of computer assisted radiology and surgery* 6 (1), 1-11 (2011).
18. Egger, J. et al. "Manual refinement system for graph-based segmentation results in the medical domain," *Journal of medical systems* 36 (5), 2829-2839 (2012).
19. Egger, J. et al. "A medical software system for volumetric analysis of cerebral pathologies in magnetic resonance imaging (MRI) data," *Journal of medical systems* 36 (4), 2097-2109 (2012).
20. Schwarzenberg, R. et al. "Cube-Cut: Vertebral Body Segmentation in MRI-Data through Cubic-Shaped Divergences," In: *PLoS One* (2014).
21. Egger, J. et al. "Template-Cut: A Pattern-Based Segmentation Paradigm," *Sci Rep.*, 2:420 (2012).
22. Egger, J. et al. "Square-cut: a segmentation algorithm on the basis of a rectangle shape," *PLoS One* 7, e31064; DOI:10.1371/journal.pone.0031064 (2012).
23. Schwarzenberg, R. et al. "A Cube-Based Approach to Segment Vertebrae in MRI-Acquisitions," *Proceedings of Bildverarbeitung für die Medizin (BVM)*, Springer Press, 69-74 (2013).

24. Egger, J., Lüddemann, T., Schwarzenberg, R., Freisleben, B. and Nimsky, C. "Interactive-cut: Real-time feedback segmentation for translational research," *Comput Med Imaging Graph.* 38(4):285-95 (2014).
25. Egger, J. et al "Interactive Volumetry of Liver Ablation Zones," *Sci. Rep.* 5:15373 (2015).
26. Egger, J. "Refinement-Cut: User-Guided Segmentation Algorithm for Translational Science," *Sci. Rep.* 4:5164 (2014).
27. Egger, J. et al. "Integration of the OpenIGTLink Network Protocol for image-guided therapy with the medical platform MeVisLab," *The International Journal of Medical Robotics and Computer Assisted Surgery*, 8(3):282-90 (2012).
28. Lu, J. et al. "Detection and visualization of endoleaks in CT data for monitoring of thoracic and abdominal aortic aneurysm stents," *Proceedings of SPIE Medical Imaging Conference*, Vol. 6918, pp. 69181F(1-7), San Diego, USA, (2008).
29. Greiner, K. et al. "Segmentation of Aortic Aneurysms in CTA Images with the Statistic Approach of the Active Appearance Models," *Proceedings of Bildverarbeitung für die Medizin (BVM)*, Berlin, Germany, Springer Press, pages 51-55 (2008).
30. Egger, J. et al. "Preoperative Measurement of Aneurysms and Stenosis and Stent-Simulation for Endovascular Treatment," *IEEE International Symposium on Biomedical Imaging: From Nano to Macro*, Washington (D.C.), USA, pp. 392-395, IEEE Press (2007).
31. Egger, J. et al. "Simulation of bifurcated stent grafts to treat abdominal aortic aneurysms (AAA)," *Proceedings of SPIE Medical Imaging Conference*, Vol. 6509, pp. 65091N(1-6), San Diego, USA (2007).
32. Xu, H. X. et al. "Treatment response evaluation with three-dimensional contrast-enhanced ultrasound for liver cancer after local therapies," *Eur J Radiol.* 76(1):81-8 (2009).